%% file: main.tex
\documentclass[10pt,twocolumn,letterpaper]{article}

\usepackage[pagenumbers]{cvpr} 

\definecolor{cvprblue}{rgb}{0.21,0.49,0.74}
\usepackage{graphicx}
\usepackage{amsmath}
\usepackage{amssymb}
\usepackage{booktabs}
\usepackage{multirow}
\usepackage{float}
\usepackage{tabularx}
\makeatletter
\@namedef{ver@everyshi.sty}{}
\makeatother
\usepackage{tikz}
\usepackage{comment}
\usepackage{color}
\usepackage{amsfonts}
\usepackage{caption}
\usepackage{mathtools}
\usepackage{tablefootnote}
\usepackage{dsfont}
\usepackage{ragged2e}
\usepackage{array}
\usepackage[pagebackref,breaklinks,colorlinks,allcolors=cvprblue]{hyperref}

\usepackage{caption}
\def\paperID{*****} 
\def\confName{CVPR}
\def\confYear{2025}
\input{sec/customized_command}
\title{DAS3R: Dynamics-Aware Gaussian Splatting for Static Scene
Reconstruction}

\author{Kai Xu$^{1}$,  \quad Tze Ho Elden Tse$^1$, \quad Jizong Peng$^2$, \quad Angela Yao$^{1}$ \\
$^1$National University of Singapore \quad
$^2$dConstruct Robotics\\
{\footnotesize \texttt{\{kxu,eldentse,ayao\}@comp.nus.edu.sg} \quad \texttt{jizong.peng@dconstruct.ai}} \\
}

\usepackage{longtable}

\usepackage{booktabs}
\usepackage{amsmath,amsfonts,amssymb}
\usepackage[nice]{nicefrac}
\usepackage{xfrac}
\usepackage{todonotes}
\usepackage{breqn}
\usepackage{soul}
\usepackage{float} 

\renewcommand{\rm}[1]{\textup{#1}}

\usepackage{amsmath}

\DeclareMathOperator*{\argmin}{arg\,min}

\newcommand{\C}[2]{\mathbf{C}^{(#1,#2)}} %
\newcommand{\X}[2]{\mathbf{X}^{(#1,#2)}} %
\newcommand{\M}[2]{\mathbf{M}^{(#1,#2)}} %

\newcommand{\G}{\mathcal{G}} %
\newcommand{\V}{\mathcal{V}} %
\newcommand{\E}{\mathcal{E}} %

\begin{document}
\twocolumn[{%
\renewcommand\twocolumn[1][]{#1}%
\maketitle
\begin{center}
\vspace{-2em}
    \captionsetup{type=figure}
    $\vcenter{\hbox{\resizebox{\linewidth}{!}{\begin{tabular}[c]{@{}c@{}}
    \includegraphics[width=\linewidth]{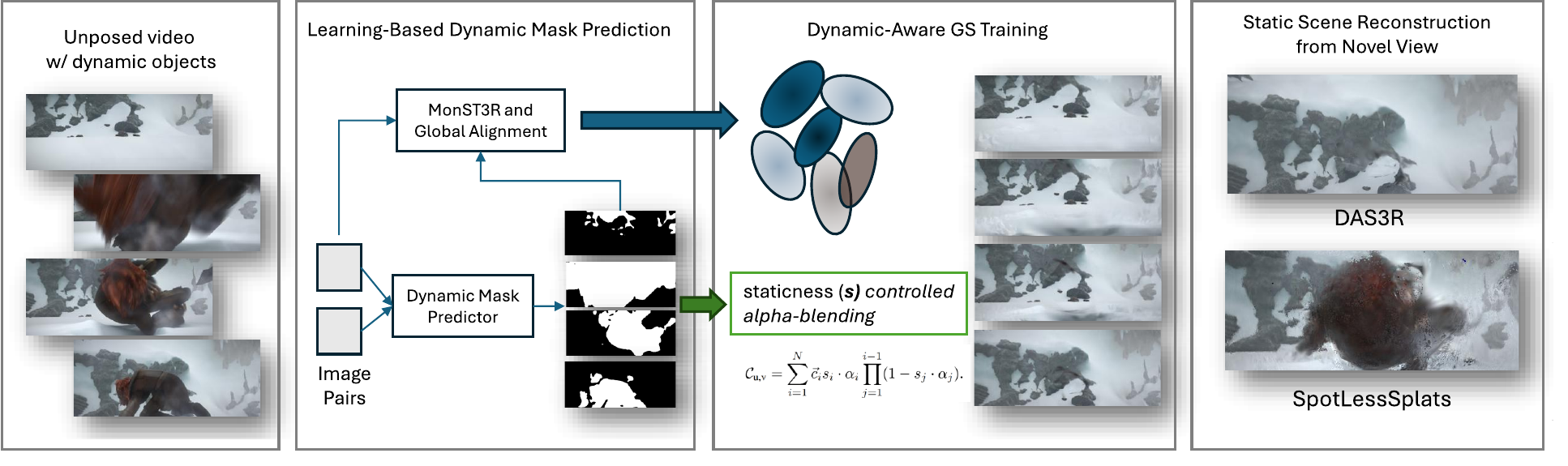} \\
    \end{tabular}}}}$
    \vspace{-0.7em}
    \captionof{figure}{
    \textbf{Overview}: DAS3R featuring reconstructing static scene from unposed videos where dynamic objects occupy a significant portion of the scene. We predict dynamic mask directly from deep network with image pair as input. The predicted dynamic masks are then used for dynamic-aware Gaussian splatting training. In the figure we show an example from Sintel dataset. Compares to SpotLessSplats \cite{sabour2024spotlesssplats}, DAS3R can reconstruct clean background while SpotLessSplats fails to remove the dynamic object. 
    }
    \label{fig:main}
\end{center}
}]
\maketitle

\input{sec/1_abstract}

\input{sec/2_intro}
\input{sec/3_related}
\input{sec/4_method}

\input{sec/5_experiment}

\input{sec/6_conclusion}
\clearpage
{
    \small
    \bibliographystyle{ieeenat_fullname}
    \bibliography{main}
}

\appendix





\end{document}

%% file: sec/customized_command.tex
\newcommand{\aftertab}{\vspace{-0.5em}}

\newcommand{\aftertabcaption}{\vspace{0.5em}}

\makeatletter
\DeclareRobustCommand\onedot{\futurelet\@let@token\@onedot}
\def\@onedot{\ifx\@let@token.\else.\null\fi\xspace}

\def\eg{\emph{e.g}\onedot} 
\def\ie{\emph{i.e}\onedot}

\makeatother

\newcommand{\ourmethod}{DAS3R}
\newcommand{\monster}{MonST3R}
\newcommand{\duster}{DUSt3R}

%% file: sec/1_abstract.tex
\begin{abstract}

We propose a novel framework for scene decomposition and static background reconstruction from everyday videos. By integrating the trained motion masks and modeling the static scene as Gaussian splats with dynamics-aware optimization, our method achieves more accurate background reconstruction results than previous works. Our proposed method is termed DAS3R, an abbreviation for \textbf{D}ynamics-\textbf{A}ware Gaussian \textbf{S}platting for \textbf{S}tatic \textbf{S}cene \textbf{R}econstruction. Compared to existing methods, DAS3R is more robust in complex motion scenarios, capable of handling videos where dynamic objects occupy a significant portion of the scene, and does not require camera pose inputs or point cloud data from SLAM-based methods. We compared DAS3R against recent distractor-free approaches on the DAVIS and Sintel datasets; DAS3R demonstrates enhanced performance and robustness with a margin of more than 2 dB in PSNR.  The project's webpage can be accessed via \url{https://kai422.github.io/DAS3R/}

\end{abstract}

%% file: sec/2_intro.tex
\section{Introduction}
\label{sec:intro}

Reconstructing static scenes from dynamic environments is a fundamental task in computer vision, with wide-ranging applications in robotic perception, animation, and the metaverse. Current methods mainly focus on autonomous driving videos \cite{li2024memorize, otonari2024entity}, or unconstrained images with transient dynamic objects \cite{sabour2024spotlesssplats, ungermann2024robust, kulhanek2024wildgaussians}, where dynamic elements typically occupy a small portion of the scene. 
These methods struggle when applied to a broader range of everyday video content that involves continuous and substantial dynamic components as well as complex camera movements. Reconstructing static background from unposed everyday video presents multiple challenges. Primarily, the movement of large dynamic objects can influence the precision of camera pose estimation.  Some methods \cite{bescos2018dynaslam, runz2018maskfusion, zhang2020vdo, xiao2019dynamic, ungermann2024robust} use segmentation networks to identify and exclude objects in video frames, but these methods often produce high false-positive rates, as semantically meaningful segmented objects are not necessarily dynamic. Other methods, such as NeRF-based or Gaussian Splatting-based dynamic scenes reconstruction \cite{DBLP:conf/cvpr/0001GMTS0C0023/RoDynRF, DBLP:journals/corr/abs-2405-17421/MoSca}, rely on thresholding epipolar errors to distinguish static and dynamic regions, with camera poses computed by robust SLAM \cite{wang2022drg} and RANSAC \cite{DBLP:journals/cacm/FischlerB81/RANSAC} for statistical outlier rejection. This method proves inadequate when a significant portion of the scene consists of dynamic objects, resulting in the statistical majority shifting from the static background to dynamic content. By integrating depth data, \cite{zhang2024monst3r} calculates an error map by contrasting the derived optical flow with the induced flow deduced from estimated camera poses, generating a dynamic mask by applying a threshold to this error. However, their performance significantly relies on accurate optical flow estimation and shows high sensitivity to the selection of threshold parameters.

We propose a direct learning-based approach to handle dynamic objects in scenes. By taking two frames of the same scene captured at different times as input and adopting \monster{} \cite{zhang2024monst3r} as the backbone network, the dynamic mask can be accurately predicted even if it occupies a significant portion of the scene.The pairwise masks are subsequently aggregated into video-level dynamic masks by combining predictions from all chosen image pairs. The primary insight is to predict a dynamic mask from an image pair instead of relying on a single image. By employing comparison and cross-attention between the two images, the underlying camera poses are decomposed from object motion in the latent space, resulting in enhanced robustness in dynamic prediction. Our method uses ground-truth dynamic object masks to supervise the learning process with the synthetic dataset Point Odyssey \cite{zheng2023point}, and is capable of generalizing to unseen datasets such as Sintel \cite{sintel} and DAVIS \cite{davis}. We follow the same post-processing pipeline of \monster{} and \duster{} \cite{DBLP:conf/cvpr/Wang0CCR24/DUSt3R} to generate dense prediction and camera pose estimation of the video. By replacing the static confidence of \monster{} in the global alignment pipeline with model predicted dynamic masks, we achieve enhanced accuracy in scene depth reconstruction and camera pose estimation. Dense predictions are subsequently employed to initialize Gaussian splats, while the predicted dynamic masks drive the dynamic-aware training for Gaussian splatting.

To achieve static background reconstruction, current efforts in robust and distractor-free reconstruction often rely on semantic  features and segmentation \cite{kulhanek2024wildgaussians, sabour2024spotlesssplats, lin2024hybridgs} and perform an implicit way to identify the transient components. While our method incorporated model predicted dynamic mask as the prior for decomposing the scene. This helps our framework to handle the persistent dynamics; we  further incorporate Gaussian Splatting training with optimizable dynamic-aware attribute - the staticness. During the rendering stage, staticness directly contributes to the computation of Gaussian point colors through alpha blending. This allows efficient removal of dynamic elements and acts as a robust solution for mitigating false-positive predictions in dynamic mask assessments.

We compared our method against recent distractor-free reconstruction methods on the DAVIS and Sintel datasets; our approach achieves a performance improvement of over 2 dB on PSNR, exhibiting strong reconstruction capabilities even with complex camera movements and substantial dynamic elements. We summarize our contributions as follows:  
\begin{itemize}  
    \item  DAS3R can predict more accurate dynamic masks from image pairs compared to the static confidence offered by previous methods.
    \item Integrating staticness as a Gaussian point attribute, DAS3R achieves effective dynamic object suppression, and accurate reconstruction under complex camera movements and substantial dynamic elements. 
    \item Detailed experimental results on DAVIS and Sintel datasets shows enhanced performance compared to distractor-free Gaussian Splatting approaches.  
\end{itemize}

%% file: sec/3_related.tex
\section{Related works}

\noindent\textbf{Dynamic Scene Decomposition.}
Estimating poses and reconstructing in dynamic settings often hinge on the segmentation of dynamic objects. Certain methods \cite{DBLP:conf/cvpr/0001GMTS0C0023/RoDynRF, DBLP:journals/corr/abs-2405-17421/MoSca} apply epipolar constraints to identify dynamic regions, relying on RANSAC for essential matrix estimation, which typically demands highly accurate correspondences. When the optical flow presented is imprecise, these methods find it challenging to decide whether the residual error arises from object movement or from faulty correspondences. 
Different approaches \cite{bescos2018dynaslam, runz2018maskfusion, zhang2020vdo, xiao2019dynamic, ungermann2024robust} emphasize the tracking and semantic segmentation of dynamic objects, generally relying on assumptions like the motion of foreground objects versus a static background, or the prior identification of mobile objects. However, these assumptions limit their applicability.\cite{akhter2008nonrigid} An alternative approach presumes that both motion segmentation and reconstruction can be concurrently addressed using factorization methods like singular value decomposition (SVD), which separate camera and object motions. Regrettably, these approaches are frequently susceptible to noise and outliers.
Our method decomposes dynamic scenes by segmentation but offers enhanced robustness and generalization by harnessing substantial volumes of training data. This allows us to reduce sensitivity to noise and outliers, thereby delivering a more scalable approach to comprehending dynamic scenes.

\noindent\textbf{End-to-end Reasoning for 3D geometry.}
End-to-end reasoning aims to build a differentiable Structure-from-Motion (SfM) pipeline to replicate traditional workflows while enabling end-to-end training~\cite{DBLP:conf/cvpr/UmmenhoferZUMID17/DeMoN, DBLP:conf/iclr/TeedD20/DeepV2D, DBLP:conf/cvpr/WangK0N24/VGGSfM}. They leverage image pairs, which is the minimal input set required for multi-view geometric computation. Models such as \duster{} and \monster{} \cite{DBLP:conf/cvpr/Wang0CCR24/DUSt3R, zhang2024monst3r} are capable of directly learning 3D pointmaps without intrinsic parameters as input. \monster{} extends \duster{} to dynamic scenes, directly regressing 3D point clouds across time steps from image pairs. DAS3R relies on \monster{} to predict dynamic masks for image pairs, and construct dynamic masks for the entire video through global alignment over frame pair graph and aggregation.

\noindent\textbf{Distractor-free 3D Gaussian Splatting.}
The traditional 3DGS  approach typically assumes inter-view consistency while this assumption can be violated when dealing with dynamic objects. Recent methods have addressed the challenge by incorporating robust 3DGS training strategies to ignore transient interferences \cite{kulhanek2024wildgaussians, sabour2024spotlesssplats, lin2024hybridgs, ungermann2024robust, otonari2024entity, turki2023suds}. For instance, WildGaussians \cite{kulhanek2024wildgaussians} combines robust DINO features with an appearance modeling module to effectively handle occlusions and appearance variations in 3DGS. SpotLessSplats \cite{sabour2024spotlesssplats} leverages Stable Diffusion features and robust optimization techniques to suppress transient interferences in 3DGS. Similarly, HybridGS \cite{lin2024hybridgs} introduces a hybrid model that integrates a multi-view-consistent 3D Gaussian model with a single-view-independent 2D Gaussian model, enabling the separation of transient and static elements in a scene.

In contrast, DAS3R focuses on capturing persistent and substantial dynamic objects from video data, while earlier approaches primarily focusing on transient dynamic objects from image data. Furthermore, DAS3R does not need camera parameters as input, while previous methods often rely on camera predictions provided by COLMAP as input.

%% file: sec/4_method.tex
\section{Preliminary}
\subsection{3D Gaussian Splatting}
3D Gaussian Splatting \cite{kerbl20233d} is a framework that represents a scene as a collection of 3D Gaussians. Each Gaussian $G(\vec{x})$ is parameterized by its mean position $\vec{\mu}$, covariance matrix $\Sigma$, color $\vec{c}_{i}$, and opacity $\alpha$:

\begin{equation}
     G(\vec{x}) = \exp({-\frac {1}{2}(\vec{x}-\vec{\mu} )^{\top }\Sigma ^{-1}(\vec{x}-\vec{\mu} )}). \label {eq:gauss}
\end{equation}

The rendering process of 3DGS involves projecting 3D Gaussian primitives onto the 2D image plane. The contribution of each Gaussian to the rendered image is determined by integrating its density along the viewing direction, weighted by its opacity $\alpha$ and color attributes $\vec{c}_{i}$:

\begin{equation}
    \mathcal{C}_{\rm{u,v}} = \sum_{i=1}^{N} \vec{c}_{i} \alpha_i \prod \limits_{j=1}^{i-1} (1- \alpha_j). 
    \label{equ:rendering}
\end{equation}

where $\mathcal{C}_{\rm{u,v}}$ denotes the final rendered color at pixel coordinate $(\rm{u},\rm{v})$, $N$ is the number of all overlapping Gaussians.

The differentiable rasterization blends the contributions of all Gaussians in the scene based on their projection onto the image plane, resulting in a continuous and differentiable image representation. Through photometric loss, each Gaussian attribute can be optimized through gradient descent based on the rendered and ground truth images. 

\begin{equation}
    \mathcal{L}_{\rm{loss}} = \mathcal{L}_{\rm{pixel}} + \lambda_{\rm{ssim}} \cdot \mathcal{L}_{\rm{ssmi}},
    \label{equ: loss}
\end{equation}

Here, ${\mathcal{L}_{\rm{pixel}}}= \|\mathcal{I}- \hat{\mathcal{I}}\|_{1}$ signifies the photometric loss calculated pixel-by-pixel, and ${\mathcal{L}_{\rm{ssim}}}= \mathcal{S}_{\rm{ssim}}(\mathcal{I}, \hat{\mathcal{I}})$ corresponds to the loss based on structural similarity. Recently, MonoGS \cite{matsuki2024gaussian} has also allowed for the simultaneous refinement of camera poses and Gaussian splats.

The presence of dynamic content in a scene results in inconsistencies for Gaussian training objectives when captures are taken at different timestamps. DAS3R employs optimizable dynamic masks to explicitly decompose the loss, directing the influence from the static scene flow solely back to the Gaussian parameters.

\section{Methodology}
\subsection{Problem Definition}

Given a with a casual captured dynamic video, our objective is to reconstruct a 3D Gaussian model of the static background, which allows for rendering that background from any viewpoint. To achieve this, we intend to train a 3D Gaussian model utilizing a dataset comprising RGB videos $\mathcal{I} = \{ I_i \mid i = 1, \dots, N \}$. Notably, the dataset for training does not include individual frame camera poses, depth details, and camera intrinsic parameters.

Each frame \( I_i \) encompasses not only photometric changes due to homography transformation caused by egocentric camera movement but also includes changes from the motion of dynamic objects. A similar setting is the distractor-free Gaussian Splatting \cite{sabour2024spotlesssplats, kulhanek2024wildgaussians}, aimed at removing transient objects for the reconstruction process from unrestrained images. In contrast to their scenarios where the dynamic objects are only transient and limited in scale, our approach tackles the challenge of reconstructing static scenes from complex dynamic videos, which involve persistent and significant dynamic objects along with intricate camera movements, as seen in the DAVIS \cite{davis} and Sintel \cite{sintel}.

\subsection{Camera Pose Estimation under Dynamic Scene}

Estimating depth and camera poses from dynamic videos poses a chicken-and-egg problem. The interplay between the motion of dynamic entities and the motion of the scene is inherently entangled. For instance, when using Gaussian Splatting to reconstruct a casual video, a 2D reprojection loss can be employed to simultaneously optimize both camera poses and the appearance of dynamic objects. When back-propagating photometric error to scene appearance and camera parameters concurrently, an ambiguity emerges regarding whether the observed changes are due to camera movement or object motion.

Previous research efforts have focused on separating camera motion from object motion through the use of epipolar error  \citep{DBLP:conf/cvpr/0001GMTS0C0023/RoDynRF, DBLP:journals/corr/abs-2405-17421/MoSca} and segmentation methods based on optical flow \citep{DBLP:journals/corr/abs-2405-18426/GFlow}. The technique presented in \monster{} \cite{zhang2024monst3r} also calculates induced flow from the estimated camera motion and depth, and subsequently determines errors based on estimated optical flow. These approaches rely heavily on precise optical flow predictions and the tuning of hyperparameters for error thresholding. Moreover, some methods \cite{ungermann2024robust} utilize semantic segmentation techniques like Segment Anything \cite{kirillov2023segany}, which depend on predictions from single images under the assumption that dynamic objects form the semantic foreground, an assumption that often does not hold in real-world contexts.

\noindent\textbf{Dynamic region segmentation}
We propose a method to train dynamic segmentation masks directly from image pair inputs. Unlike traditional semantic segmentation networks that rely on a single image for predictions, our dynamic approach relies on reasoning across two overlapping image frames. In particular, we employ \monster{} as the foundational model for predicting the dynamic masks. Adopting the approach of \duster{}\cite{DBLP:conf/cvpr/Wang0CCR24/DUSt3R} and  CroCo v2 \cite{croco_v2}, \monster{} performs direct regression of dense 3D point maps from the image pairs exhibiting dynamic content, eliminating the need for prior scene or camera information, such as intrinsic parameters.

Specifically, the input consists of two RGB images with some overlapping content, \( I_n \) and \( I_m \), captured at different time points \( t_n \) and \( t_m \), where \( I_n, I_m \in \mathbb{R}^{W \times H \times 3} \). The output includes two corresponding point maps \( \X{n}{n} \) and \( \X{n}{m} \in \mathbb{R}^{W \times H \times 3} \), and associated confidence maps $\C{n}{n},\C{m}{n}$. The first subscript indicates the coordinate system of the point maps, which are both expressed in the same coordinate system as the image \( I_1 \), while the second subscript indicates the frame index of the point map.  In addition to \monster{}, we also predict a dynamic mask \(\M{n}{n} \in \mathbb{R}^{W \times H \times 1}\) that designates whether points are part of a dynamic object region. For saving computations, we only predict the dynamic mask for frame 1 as it is sufficient to be aggregated to image dynamic masks.
The dynamic mask prediction, which employs the DPT head \cite{ranftl2021vision}, utilizes intermediate features derived from Croco backbones. This includes the encoded features from each image and the cross-attention output's intermediate features. For successful training of this model, we produce ground-truth dynamic masks for Point Odyssey \cite{zheng2023point}. Point Odyssey provides sparse 3D points motion information based on its 3D trajectories, from which we can generate the image-level dynamic mask by nearest interpolation to the image pixel's 3D projection.

\begin{figure}[h]
  \centering
    \caption{Dynamic Mask Comparison on DAVIS dataset.  }\label{fig:davis_mask}
    
  \includegraphics[width=1\linewidth]{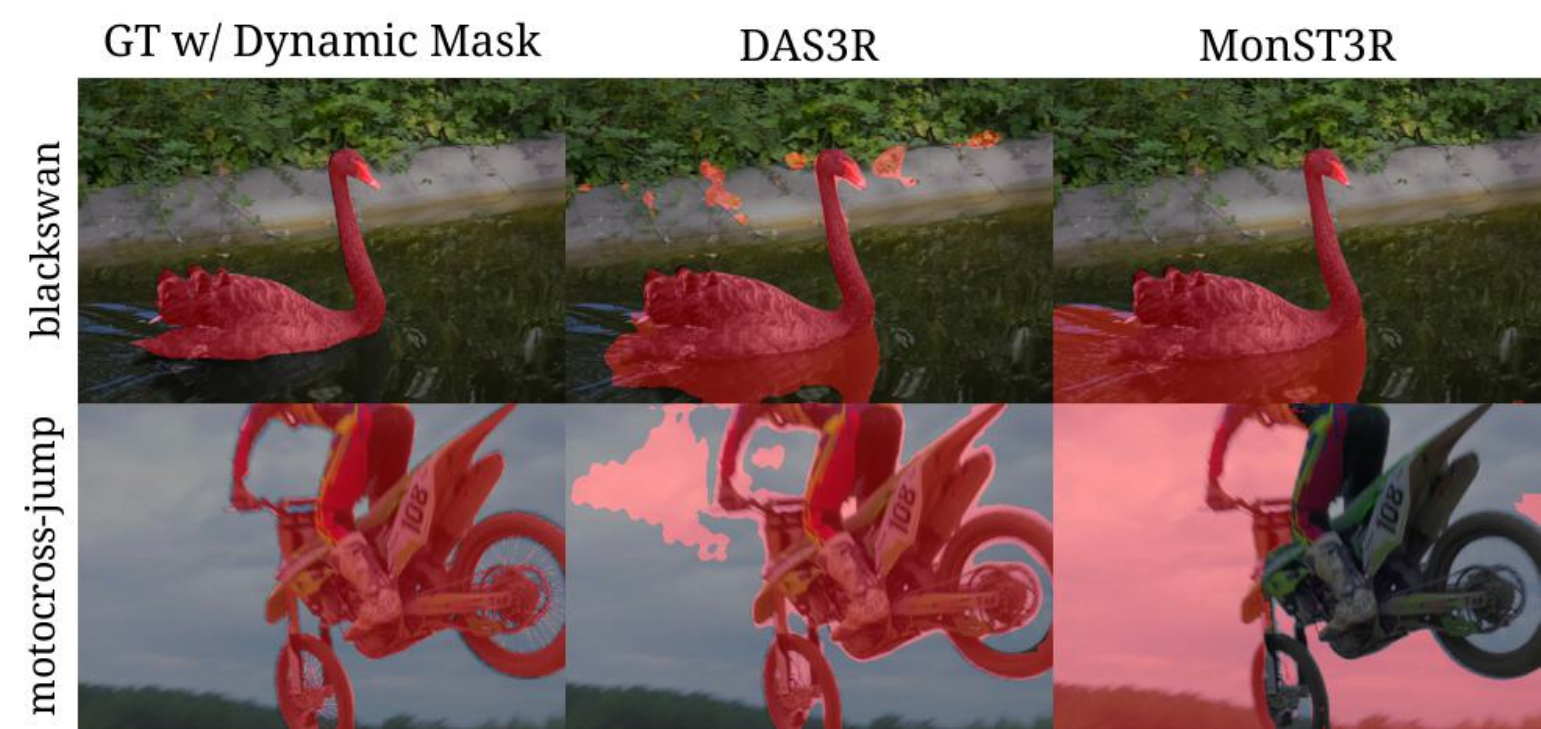}

\end{figure}

\begin{figure}[h]
  \centering
    \caption{Dynamic Mask Comparison on Sintel dataset.  }\label{fig:sintel_mask}
    
  \includegraphics[width=1\linewidth]{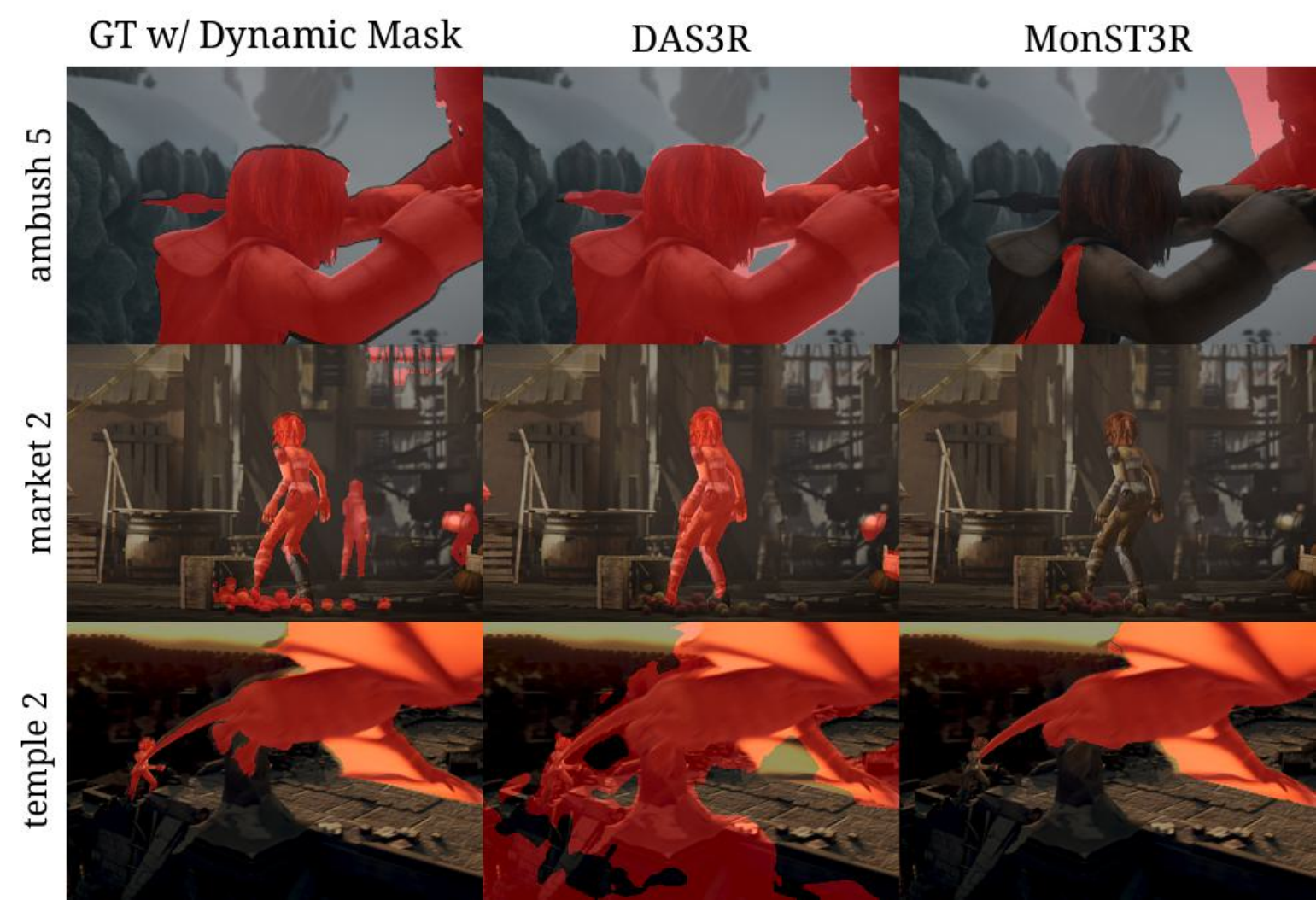}

\end{figure}

\noindent\textbf{Comparison to \monster{}}
We perform both qualitative and quantitative comparisons with static confidence from \monster{} as shown in Table \ref{tab:dynamic_mask} and Figures \ref{fig:davis_mask} and \ref{fig:sintel_mask}. Our approach delivers more accurate segmentation results, whereas \monster{} often either misses segments or incorrectly identifies static areas as dynamic.One limitation of our model's prediction mask is the occurrence of false positives in scenarios involving significant depth variation, which complicates the differentiation between static and dynamic foregrounds. To address this challenge, we integrate staticness into the Gaussian property and further refine it during training. More details are provided in Section \ref{sec:gs}.

\begin{table}[t]
\centering
\resizebox{0.5\linewidth}{!}{
\begin{tabular}{|l|c|c|}
\hline

Method & DAVIS  & Sintel    \\ 

\hline

\monster{}  &{32.5}&{37.1}  \\ 

Ours        &{39.7}&{59.3}  \\ 
\hline

\end{tabular}
}

\caption{Comparison of dynamic mask accuracy (IoU) on DAVIS and Sintel datasets.}\label{tab:dynamic_mask}
\end{table}

\subsubsection{Global Alignment with Predicted Dynamic Mask}
We follow \duster{} \cite{DBLP:conf/cvpr/Wang0CCR24/DUSt3R} and \monster{} \cite{zhang2024monst3r} in transforming pair predictions into the video's global camera poses. \duster{} introduced a global alignment loss, while \monster{} proposed an approach involving optical flow reprojection loss and smoothing loss, utilizing optimized static confidence to adjust the loss. In our approach, we replace static confidence by directly aggregating pairwise dynamic mask predictions into image dynamic masks.

Let $\G(\V,\E)$ be the connectivity graph of the frames' indices. $\V$ denotes vertices and $\E$ denotes edges, where each edge $e=(n,m) \in \E$ denotes pair images $I^n$ and $I^m$ which share similar visual content for pointmap computation. 

For each image pair $e=(n,m) \in \E$, the pairwise pointmaps $\X{n}{n}, \X{m}{n}$, associated confidence maps $\C{n}{n},\C{m}{n}$, and soft dynamic mask $\M{n}{n}$.

Let us assume the optimal global pointmap at time $t$ is $\mathbf{X}^{(t,\mathbf{w})}$, the first $t$ denotes frame index and the second $W$ denotes that the pointmap is under the world coordinates. 
for edge $(m,n)$, 
\begin{multline}
\mathcal{L}_{\textrm{align}} (\mathbf{X}^{(\cdot,\mathbf{w})}, \sigma, \G(\V,\E))  = \\
\sum_{e = (n,m) \in \E} \Bigl(|| \mathbf{C}^{(n,n)}  (\mathbf{X}^{(n,\mathbf{w})} - \sigma^{e} \mathbf{P}^{n} \mathbf{X}^{(n,n)}_e) ||_1
+ \\
|| \mathbf{C}^{(m,n)} \cdot (\mathbf{X}^{(m,\mathbf{w})} - \sigma^{e} \mathbf{P}^{n} \mathbf{X}^{(m,n)}_e)||_1 \Bigr),
\end{multline}
$\sigma^{e}$ denotes a pairwise scale factor, and $\mathbf{P}^n$ is the camera pose transformation at frame $n$. Apart from pointmap alignment loss, we also use the smoothness loss from \monster{} for encouraging smooth camera motion: 
$\mathcal{L}_{\text{smooth}}(\mathbf{X})  = 
\sum_{t=0}^{N} \left( \left\| {\mathbf{R}^t}^\intercal \mathbf{R}^{t+1} - \mathcal{I} \right\|_{\text{f}} + \left\| {\mathbf{R}^t}^\intercal (\mathbf{T}^{t+1} - \mathbf{T}^t) \right\|_2 \right),
$ where $\| \cdot \|_{\text{f}}$ denotes the Frobenius norm, and $\mathcal{I}$ is the identity matrix.

\noindent\textbf{Flow loss with predicted dynamic mask} In \monster{}, the flow projection loss aims for the global pointmaps and camera positions to align with the calculated flow concerning the confident static areas of the real frames, depending on determining static confidence regions using PnP and RANSAC. We replace the static confidence regions with our model's dynamically predicted masks.
\begin{equation}
\mathcal{L}_{\text{flow}}(\mathbf{X}) = \sum_{W^i\in W} \sum_{t\rightarrow t'\in W^i} || (1-\mathbf{M}^{t}) \cdot (\textbf{F}_{\textrm{cam}}^{t\rightarrow t'} - \textbf{F}^{t\rightarrow t'}_{\textrm{est}} )||_1,
\end{equation}
where $\textbf{F}_{\textrm{cam}}^{t\rightarrow t'}$ is the induced flow and $\textbf{F}_{\textrm{est}}^{t\rightarrow t'}$ is the estimated flow. 
$\mathbf{M}^{t}$ is the aggregated masks from all pairs including frame t. \ie
\begin{equation}
\mathbf{M}^{t} = \sum_{(t,m)\in\mathcal{E}_{t*}}  \mathbf{M}^{t,t} / |\mathcal{E}_{t*}| ,
\end{equation}
where $|\cdot|$ denotes the cardinality of the set.
The complete optimization for our dynamic global point cloud and camera poses is
 $
    \hat{\mathbf{X}}
    =\argmin_{\mathbf{X}, \mathbf{P}_W, \sigma} \mathcal{L}_{\textrm{align}}+ w_{\text{smooth}} \mathcal{L}_{\textrm{smooth}} + w_{\text{flow}} \mathcal{L}_{\textrm{flow}},
$ where $w_{\text{smooth}}, w_{\text{flow}}$ are hyperparameters. 

\noindent\textbf{Evaluation on Sintel and TUM-dynamics}
Table \ref{tab:camera_pose} presents a comparison of camera pose estimation results on the Sintel and TUM-dynamics datasets. Integrating trained dynamic masks into the optimization process significantly enhances the overall accuracy of the camera pose, with notable improvements in rotation accuracy.
\input{tab/pose_estimation}

\subsection{Static Background Reconstruction with 3D Gaussian Splatting}\label{sec:gs}

We begin initializing the Gaussian splats by utilizing prediction from global alignment, integrating both the depths and  camera parameters for each frame. Following Instant-Splat \cite{fan2024instantsplat}, the dense 3D point clouds are filtered based on the model's confidence values. Furthermore, each 3D Gaussian is linked with its accumulated dynamic probability $\mathbf{M}$, and this is added as an attribute of the Gaussian splats.

Specifically, \textbf{staticness} (\( s \)) is defined as the complement of the dynamic mask probability (\( P_{\text{dynamic}} \)):

\[
s_{u,v,t} = 1 - \mathbf{M}^t_{u,v}, \in [0, 1]
\]

where $u,v$ denotes the pixel index of the initialization.

To address errors and noise in the predicted, especially false-positive segmentations in scenes with significant depth variation, we optimize the staticness property as an attribute of the Gaussian primitives. Each Gaussian is assigned an attribute called \textbf{staticness} (\( s \)), which quantifies the likelihood that the Gaussian primitive belongs to the static region. During rendering, the staticness is incorporated into the alpha blending computation to filter out dynamic objects when reconstructing static scenes. 

The final alpha value for rendering is computed by:
\begin{equation}
    \mathcal{C}_{\rm{u,v}} = \sum_{i=1}^{N} \vec{c}_{i} s_i\cdot\alpha_i \prod \limits_{j=1}^{i-1} (1- s_j\cdot\alpha_j). 
    \label{equ:rendering}
\end{equation}

Incorporating staticness into the Gaussian representation, \ourmethod{} successfully diminishes the impact of dynamic objects and corrects false-positive predictions of dynamic masks, improving the reconstruction quality of static scenes.

\noindent\textbf{Static alignment for camera poses refinement.} 
In Section 4.2, we presented the loss functions used by \monster{} for achieving global alignment. These losses arise from analyzing local 3D point clouds constructed from depth information in each frame, with optimization performed on an individual frame level rather than across the full global point cloud space of the video. Nonetheless, aligning within the global 3D space poses two significant challenges: 1) the 3D point clouds for individual frames are discrete and created by reprojecting pixels, thus complicating accurate alignment, and 2) the inherently dynamic nature of the content opposes alignment.

To achieve global alignment of 3D geometries, we employ 3D Gaussians as the global representation for the entire scene. By aligning the static components of the scene, the accuracy of camera pose estimation can be enhanced, leading to improved reconstruction quality. The alignment loss is defined as the \( L_1 \) loss on the rendered image, calculating the pixel-wise discrepancy between the rendered image \( \mathbf{I}_{\text{rendered}} \) and the actual image \( \mathbf{I}_{\text{gt}} \).
\begin{equation}
    \mathcal{L}_{\text{image}} = \frac{1}{|\Omega|} \sum_{\mathbf{p} \in \Omega} \left\lVert \mathbf{I}_{\text{rendered}}(\mathbf{p}) - \mathbf{S(p)}\cdot\mathbf{I}_{\text{gt}}(\mathbf{p}) \right\rVert_1,
\end{equation}

where $\mathbf{S(p)}$ denotes per-frame Staticness. \( \Omega \) is the set of all pixels in the image, and \( \mathbf{p} \) is the 2D pixel coordinate.

Throughout the optimization, gradients are back-propagated to the camera pose parameters (\( \mathbf{R}_i \) for rotation and \( \mathbf{t}_i \) for translation) by re-projecting observed points into the Gaussian model. 

\noindent\textbf{Training.}
During the Gaussian optimization process, we allow only the position and opacity of the Gaussians to be optimized, while keeping their color parameters, rotation, and scale fixed. The opacities are initially set to $1/N$, with $N$ representing the total number of frames, to ensure that gradients are propagated across all points in the scene. Since a sufficient number of initialization points have already been obtained during the initialization phase, we disable further cloning of Gaussians. Table \ref{tab:training_cost} demonstrates that this strategy effectively accelerates the convergence of camera pose predictions while already achieving photometric performance that surpasses the baseline.

%% file: tab/pose_estimation.tex
\begin{table}[th]
\centering
\footnotesize
\renewcommand{\arraystretch}{1.3}
\renewcommand{\tabcolsep}{2.5pt}
\caption{Evaluation of camera pose estimation on Sintel and TUM-dynamic datasets.}
\aftertabcaption
\label{tab:camera_pose}
\resizebox{\linewidth}{!}{
\begin{tabular}{lccc|ccc}
\hline
& \multicolumn{3}{c}{Sintel} & \multicolumn{3}{c}{TUM-dynamics}   \\ 
\cline{2-4} \cline{5-7} 
{Method} & {ATE $\downarrow$} & {RPE trans $\downarrow$} & {RPE rot $\downarrow$} & {ATE $\downarrow$} & {RPE trans $\downarrow$} & {RPE rot $\downarrow$}   \\ 
\hline
Robust-CVD & 0.360 & 0.154 & 3.443 & 0.153 & 0.026 & 3.528  \\ 
CasualSAM & 0.141 & {0.035} & {0.615} & 0.071 & 0.010 & 1.712  \\ 
\monster{}* & {0.108} & {0.042} & {0.732} & {{0.104}} & {{0.022}} & {{1.042}}  \\ 
\hline
\textbf{Ours} & {{0.107}} & {0.041} & {0.669} & {{0.072}} & {{0.019}} & {{0.948}}  \\ 
\hline
\addlinespace[2pt]
\multicolumn{7}{l}{
     $^*$ The results for \monster{} on TUM-dynamics are reproduced by us.}
\end{tabular}
}
\aftertab
\end{table}

%% file: sec/5_experiment.tex
\section{Experiment}

\subsection{Experimental Details}

\noindent \textbf{Training of dynamic mask network.} 
We utilize the pre-trained model of \duster, keeping the ViT encoder and decoder fixed, while training the DPT head for dynamic mask prediction network. Following the approach of \monster, we train for a total of 50 epochs, using 20,000 sampled image pairs per epoch. We employ the AdamW optimizer with a learning rate of \(5 \times 10^{-5}\) and a mini-batch size of 4 per GPU. For the dataset, we use Point Odyssey \citep{zheng2023point}, which is a dataset of 200k synthetic images that feature 131 realistic indoor and outdoor scenes.

We utilize the trajectories of ground truth points (which are sparse) to calculate the ground-truth dynamic mask. Since the points generated by the network are inverse-projected from the 2D pixel image onto 3D space, they do not completely align with the ground truth points. We apply nearest interpolation using points that possess dynamic indicators.

\subsection{Static Scene Reconstruction}

We use the Sintel \cite{sintel} and DAVIS \cite{davis} datasets as the benchmark for our experiments. The Sintel dataset is a synthetic dataset which is well-known for its dynamic complexity, featuring challenging scenes with intricate motion patterns and interactions between dynamic and static elements, making it ideal for evaluating dynamic video decomposition and reconstruction methods. The DAVIS dataset is a benchmark dataset widely used for video object segmentation tasks. From DAVIS 2016, we selected the initial 50 frames from 8 sequences wherein the ground-truth masks exclusively delineate the moving objects.

In our experiments, we report PSNR for novel static scene reconstruction. Since the Sintel and DAVIS datasets do not provide ground-truth backgrounds, we exclude dynamic objects from the PSNR calculation by applying the ground-truth dynamic mask. In the case of Sintel, these dynamic masks are generated using ground-truth camera poses and optical flow information. We implement a 90-10 split: 90\% of the frames are designated for Gaussian model training, and the remaining 10\% are designated as test frames for assessing PSNR. Additionally, during training, we refine the camera poses of test frames to ensure precise rendering during evaluation.

\subsection{Static Scene Reconstruction}

\noindent\textbf{Comparison on DAVIS and Sintel Dataset} We compared the distractor-free Gaussian reconstruction methods employed in unconstrained image sets, including WildGaussians \cite{kulhanek2024wildgaussians}, Robust3DGS \cite{ungermann2024robust}, and SpotLessSplats \cite{sabour2024spotlesssplats}, as well as a pose-free baseline method combining \monster{} \cite{zhang2024monst3r} and InstantSplat \cite{fan2024instantsplat}. The camera parameters calculated by COLMAP \cite{schoenberger2016mvs, schoenberger2016sfm} were used to evaluate the results of the first three methods. 

\begin{figure*}[htbp]
  \centering
  \caption{Qualitative comparison on DAVIS dataset. \ourmethod{} achieves best rendering quality and is able to correctly detect and remove dynamic objects with their shadow and reflection. SpotLessSplats removes static content (the hurdles) in horsejump-high (4th row of DAVIS).
}\label{fig:comparison_davis}
    
  \includegraphics[width=1\textwidth]{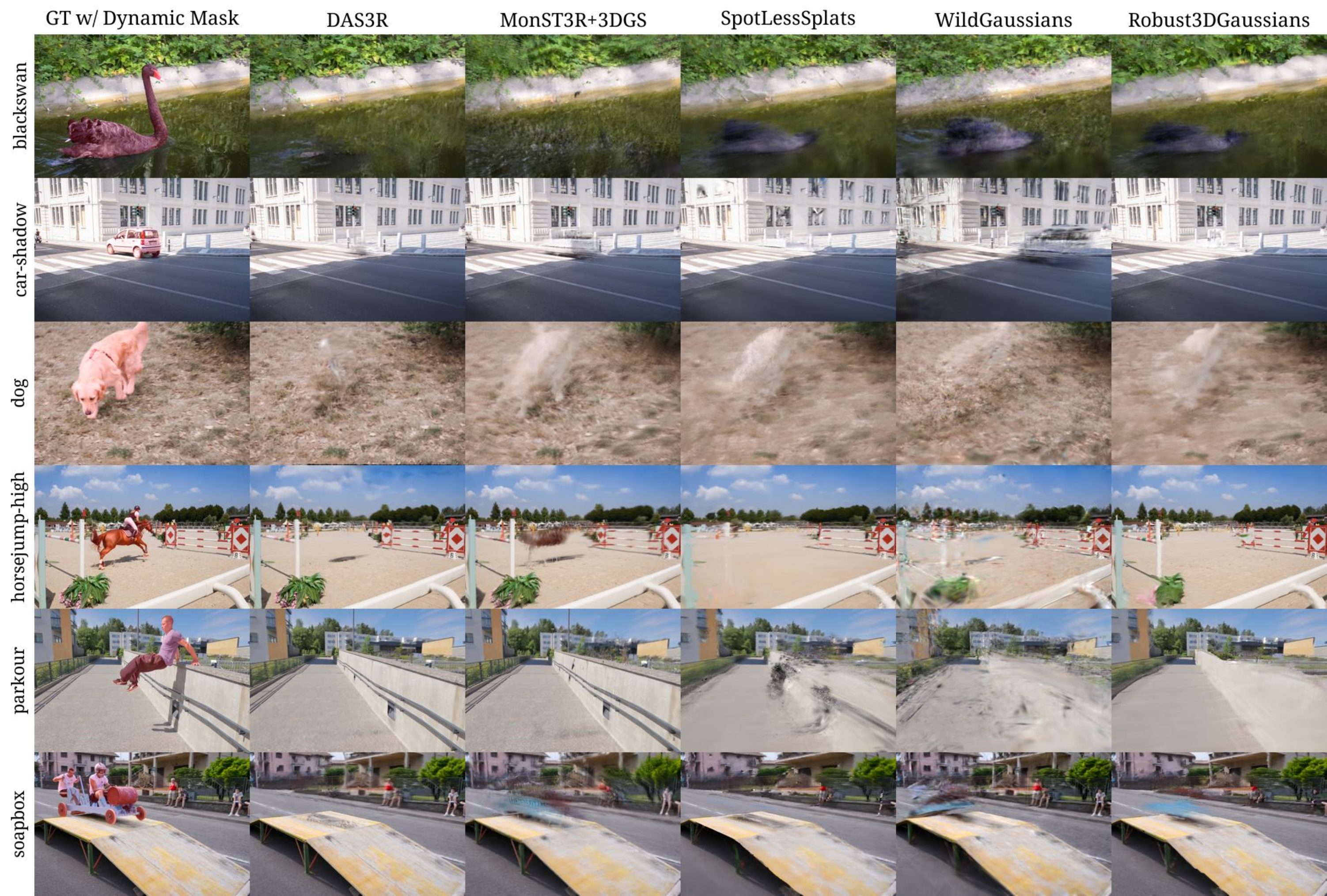}
\end{figure*}

\begin{figure*}[h!]
  \centering
    \caption{Qualitative comparison on Sintel dataset. \ourmethod{} is robust to large dynamic objects while other methods fail to remove the dynamic objects and even fail to reconstruct the overall scene . }\label{fig:comparison_sintel}
    
  \includegraphics[width=1\textwidth]{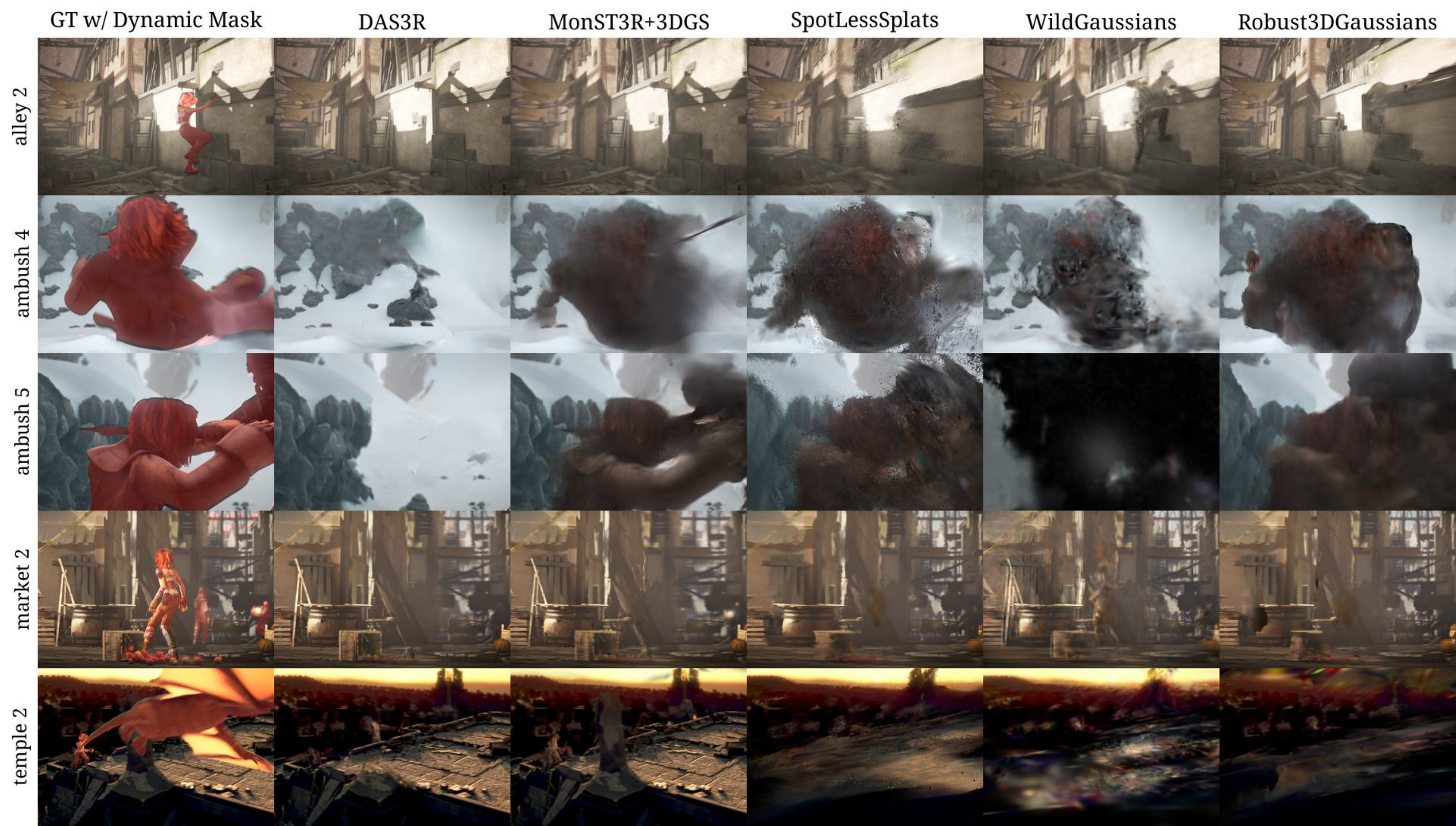}

\end{figure*}

We provide quantitative comparisons for DAVIS dataset in Table \ref{tab:psnr_comparison_davis} and Sintel dataset in Table \ref{tab:psnr_comparison_sintel}.
\ourmethod{} obtains highest average scores even without any camera poses as input. We also provide qualitative comparisons for DAVIS dataset in Figure \ref{fig:comparison_davis} and Sintel dataset in Figure \ref{fig:comparison_sintel}. WildGaussians fails to reconstruct when camera movement is large and a dynamic object takes large camera space, \eg ambush-5 (2nd row of Sintel) and ambush-6 (3rd row of Sintel). Robust3DGS and SpotLessSplats fail to remove dynamic objects once they take a lot of camera space. For some cases, \eg horsejump-high (4th row of DAVIS), SpotLessSplats removes a lot of static content (the hurdles ). Our \ourmethod{} is able to correctly detect dynamic objects (even with their shadow and reflection) and remove them accordingly.

\begin{table*}[h!]
\centering

\resizebox{\linewidth}{!}{ 
\begin{tabular}{l|cccccccc|c}
& blackswan & camel & car-shadow & dog & horsejump-high & motocross-jump & parkour & soapbox & Average \\ 

\hline
WildGaussians & 18.95 & 19.19 & 21.45 & 19.74 & 18.79 & 7.91 & 18.89 & 20.55 & 18.18 \\
Robust3DGS & 19.58 & 21.31 & 29.31 & 22.48 & 20.87 & 13.83 & 21.29 & 22.55 & 21.40 \\
SLS-mlp &21.14 & 25.62 & 22.77 & 23.82 & 18.78 & 17.82 & 23.15 & 22.43 & 21.94\\
\hline
\monster{} + InstantSplat & 20.30 & 20.97 & 25.55 & 24.41 & 24.38 & \textbf{18.95} & 25.26 & 25.35 & 23.14\\
\ourmethod{} w/o static conf & \textbf{24.12} & 27.06 & \textbf{31.04} & 28.53 & 21.11 & 17.92 & 26.90 & 26.11 & 25.35\\
\ourmethod{} & 23.90 & \textbf{27.27} & 29.13 & \textbf{28.63} & \textbf{25.09} & 17.09 & \textbf{28.09} & \textbf{26.41} & \textbf{25.70} \\ 
\end{tabular}
} 

\caption{Comparison on DAVIS dataset. The PSNR is computed on static area by masking out dynamic content.  The best results are \textbf{bold}. \ourmethod{} achieves best average results, even without COLMAP poses given.}\label{tab:psnr_comparison_davis}
\end{table*}

\begin{table*}[htbp]
\centering

\resizebox{\linewidth}{!}{ 
\begin{tabular}{l|cccccccccccccc|c}
& alley-2 & ambush-4 & ambush-5 & ambush-6 & cave-2 & cave-4 & market-2 & market-5 & market-6 & shaman-3 & sleeping-1 & sleeping-2 & temple-2 & temple-3 & Average\\ \hline

WildGaussians & 16.75 & 21.43 & 7.87 & 4.02 & 26.69 & 27.68 & 23.14 & 11.47 & 16.56 & 32.29 & 15.38 & 17.06 & 16.50 & 15.15 & 18.00\\

Robust3DGS & 17.96 & 19.18 & 12.20 & 10.46 & 27.70 & 29.68 & 22.28 & 16.85 & 16.23 & 35.88 & 15.58 & 15.93 & 12.68 & 20.68 & 19.52\\
SLS-mlp & 19.09 & 19.75 & 14.14 & 5.50 & 27.62 & 29.20 & 23.74 & 17.73 & 17.76 & 36.84 & 19.05 & 21.61 & 19.12 & 22.12 & 20.95\\
\hline
\monster{}+InstantSplat & 27.70 & 22.51 & 18.04 & 14.75 & 27.40 & 31.94 & 27.12 & 23.57 & \textbf{26.86} & 43.89 & \textbf{31.20} & \textbf{35.73} & \textbf{28.84} & 21.00 & 27.18 \\
\ourmethod{} w/o static conf & \textbf{31.50} & \textbf{24.61} & 26.23 & 18.71 & 28.32 & \textbf{32.06} & 28.11 & 26.41 & 20.05 & 44.45 & 14.98 & 14.70 & 21.44 & 23.78 & 25.38\\
\ourmethod{}  & 31.10 & 24.52 & \textbf{26.28} & \textbf{19.26} & \textbf{28.32} & 31.86 & \textbf{29.03} & \textbf{26.49} & 23.58 & \textbf{45.60} & 26.30 & 25.67 & 27.18 & \textbf{23.90} & \textbf{27.79} \\ 
\end{tabular}
} 
\caption{Comparison on Sintel dataset. The PSNR is computed on static area by masking out dynamic content.  The best results are \textbf{bold}. \ourmethod{} achieves best average results, even without COLMAP poses given.}\label{tab:psnr_comparison_sintel}
\end{table*}

\noindent\textbf{Gaussian Splatting training efficiency} We also compare the cost of training in Table \ref{tab:training_cost}. \ourmethod{} only requires 4000 iterations, while the other methods require at least 30000 iterations, which take 10 minutes to train on a video of length 50. 

\begin{table}[htbp]
\centering

\resizebox{0.7\linewidth}{!}{ 
\begin{tabular}{l|cc}
& Iterations & Time\\ \hline
\ourmethod{} & 4000 & $\sim$2mins\\
WildGaussians & 70000 & $\sim$40mins\\
Robust3DGS & 30000 & $\sim$10mins\\
SLS-mlp & 30000 & $\sim$10mins\\
\end{tabular}
} 
\caption{Comparison on training cost. Test on RXT 4090 on a 480p video of 50 frames.}\label{tab:training_cost}
\end{table}

\subsection{Dynamic Mask Accuracy}

Table \ref{tab:dynamic_mask} presents the accuracy results for the dynamic masks on DAVIS and Sintel, evaluated using IoU accuracy.
Figure \ref{fig:davis_mask} and Figure \ref{fig:sintel_mask} show some examples comparing \ourmethod{} and \monster{}'s dynamic mask. Our methods provide more valid masks than \monster{} when there are large dynamic objects. One of the limitations is our method tends to predict false positives on static areas; this could be mitigated by Staticness optimization during Gaussian Splatting training.

\subsection{Camera Pose Accuracy}
We provide camera pose estimation accuracy from the global alignment step in Table \ref{tab:camera_pose}. Our method achieves remarkable improvement in Relative Rotation accuracy on Sintel, and all metrics on TUM-dynamics. This result demonstrates the necessity of utilizing more accurate dynamic masks for global camera pose alignment. Better camera poses also provide initialization for training Gaussian Splats.

\subsection{Ablation on Staticness}

We verify the effectiveness of Staticness in terms of rendering accuracy of trained Gaussian Splats. Results are provided in Table \ref{tab:psnr_comparison_davis} and Table \ref{fig:comparison_sintel}. Disabling Staticness leads to degraded results, especially on hard samples where the network falsely predicts the static content as dynamic objects.

%% file: sec/6_conclusion.tex
\section{Conclusion}
This work introduces DAS3R, a novel framework for scene decomposition and static background reconstruction from dynamic monocular videos. By leveraging dynamics-aware Gaussian splatting and accurately predicting dynamic masks through a learning-based approach, DAS3R achieves robust and effective reconstruction in challenging scenarios involving significant dynamic objects and complex camera movements. Compared to prior methods, DAS3R demonstrates superior performance, evidenced by substantial gains in PSNR and enhanced camera pose estimation across benchmark datasets like DAVIS and Sintel.
Despite its robustness, the system occasionally suffers from false positives in dynamic object detection, particularly in areas with significant depth variation. Addressing these limitations through more diverse training data and improved model refinements will be the focus of future work.